# A Fingerprint-based Access Control using Principal Component Analysis and Edge Detection

Ernande F. Melo [#1], H. M. de Oliveira [*2]

[#]*Computer Engineering Department, Amazon State University*
Av. Darcy Vargas, 1200, Parque 10, 69065-020, Manaus, Amazonas, Brazil

[#]*Superior Teaching Center, Foundation Center of Analisys, Research and Tec. Innovation-FUCAPI*
Av. Danilo Areosa, 381, DI, 69074-130, Manaus, Amazonas, Brazil
[1] `ernande.melo@fucapi.br`

[*]*Signal Processing Group, Federal University of Pernambuco*
C.P. 7800, 50711-970, Recife, Brazil
[2] `hmo@ufpe.br`

*Abstract*— **This paper presents a novel approach for deciding on the appropriateness or not of an acquired fingerprint image into a given database. The process begins with the assembly of a training base in an image space constructed by combining Principal Component Analysis (PCA) and edge detection. Then, the parameter value ($H$) – a new feature that helps in the decision making about the relevance of a fingerprint image in databases – is derived from a relationship between Euclidean and Mahalanobian distances. This procedure ends with the lifting of the curve of the Receiver Operating Characteristic (ROC), where the thresholds defined on the parameter $H$ are chosen according to the acceptable rates of false positives and false negatives.**

*Keywords*- **Fingerprints, PCA, Edge Detection, Euclidean and Mahalanobian distances, ROC.**

## I. INTRODUCTION

Fingerprints are currently emerging as the widespread biomedical identification method in both forensic and civilian applications. Different issues related to fingerprint classification have been addressed [1-3]. Nowadays, many automatic fingerprint identification systems (AFIS) are available. There are an astonish number of sophisticated techniques in the scope of fingerprint classification and identification. Tools such as FFT [3], wavelets [4], neural network [5], Principal Component Analysis [7-9] have been suggested. But they have long been mainly used in two scenarios, namely authentication (Am I who I claim?) and recognition (who am I?). This paper is concerned with the former topic. It is essentially concerned with verification, introducing a straightforward fingerprint-based access control algorithm. The central question is whether the holder of a fingerprint may be allowed to enter a facility (e.g. a particular Lab at the university). It is somewhat equivalent to correctly deciding whether a fingerprint image is within the system database or not. We consider four possible answers: (i) the acquired fingerprint image is found to be into the base when actually it is not in (miss – false positive); (ii) the fingerprint image is not recognize as being in the base, but it is actually in the base (miss – false negative); (iii) the fingerprint is found to be in the base and it actually is (hit); (iv) the acquired fingerprint is not recognized as being in the image base and it is not in it (hit). The performance evaluation of the categorizing scheme to verify the pertinence of a fingerprint image in the given data base is essentially related with the minimization of the number of false positives (FP) and false negatives (FN). This article presents a novel procedure whose performance is analyzed in terms of the Receiver Operating Characteristic curve (ROC) [6]. Unlike further approaches [1, 2, 5], the development this new algorithm does not exploits directly the characteristics inherent to the geometry of a fingerprint, but rather uses the approach "put all together", i.e. in order to settle on a fingerprint, it is applied a set of transformations such as Principal Component Analysis (PCA) [7, 9], Edge Detection [10], Euclidean distances [11], and Mahalanobian distance [11]. The discrepancy that accounts for the fine performance of this procedure in a number of noisy scenarios (Section V) is the insertion of the edge detection feature combined with the definition of a new parameter $H$. This parameter allows a more clear-cut selection of decision thresholds about the (possible) pertinence of the acquired fingerprint image in a stored database.

## II. A FINGERPRINT CLASSIFICATION ALGORITHM

In this section, we introduce a basic version of the algorithm for fingerprint classifying (FPC), which has as preliminary input a database of fingerprint images (*Db_img*). A test fingerprint image ($I_X$) is then entered, and the algorithm returns whether or not the test image is in the stored fingerprint bank.

*A. Description of the algorithm FPC at the first level*

Input parameters (*Db_img*, $I_X$).
Step1: Store previously the fingerprint database denoted by *Db_img* (Fingerprint databank);
Step2: Build the image space by using the PCA technique;

Step3: Acquire and project the fingerprint testing image ($I_X$) into image space;
Step4: Define a criterion for classifying the ($I_X$);
Step5: Decide whether $I_X$ belongs or not to the stored base.

### B. Refining Step 2

Building the image space by using the PCA technique; The PCA is a technique that allows, from extracting the eigenvectors and eigenvalues of the covariance matrix of images, to create a space of reduced images, which contains the "main components" of the fingerprint images for subsequent recognition [7-9]. Although widespread adopted in the recognition of face images [8-9], we apply here this technique as a part of the fingerprints classification algorithm. In the sequel we illustrate the technique of PCA applied in this FPC algorithm:

1) Collect a set of $M$ input images ($I_m$) each of size ($N \times K$), Transform such images in a single vector and store in the form ($N.K \times 1$), thus generating the array $Db\_img$ ($N.K \times M$) as follows:

$$I_m = \begin{bmatrix} p_{11} & p_{21} & p_{1K} \\ p_{21} & p_{22} & p_{2K} \\ & \ddots & \\ p_{N1} & p_{N2} & \cdots p_{NK} \end{bmatrix} := \begin{bmatrix} P_1 \\ P_2 \\ \vdots \\ P_N \end{bmatrix}_{N \times K} \rightarrow$$

$$I_m = stack(P_1, P_2, \ldots P_N)_{NK \times 1} \rightarrow Db\_img = [I_1^T : I_2^T : \ldots : I_M^T]_{NK \times M}$$

2) Compute the mean of the images ($Imed$) and then the difference between $Imed$ and each image in $Db\_img$, generating thus the $A$ matrix, calculated as follows;

$$I_{med} := \frac{1}{M} \sum_{m=1}^{M} I_m = \begin{bmatrix} Im_1 \\ Im_2 \\ \vdots \\ Im_{NK} \end{bmatrix}_{NK \times 1} \rightarrow$$

$$A := [Db\_img(-)(I_{med})]_{NK \times M}$$

3) Compute $K_A$, the covariance matrix of $A$, i.e. $K_A := AA^T$ which results in the dimensions: ($NK \times M$)*($M \times NK$)=$NK \times NK$;

4) Find out the eigenvectors and eigenvalues of the covariance matrix $K_A$. Equipped with these eigenvectors, we construct the space of images containing the most relevant components of the fingerprints for the classification process. However it is well known that the finding of eigenvectors in an array of dimension $NK \times NK$ is a computational intractable task for the image sizes most frequently found [8, 9]. A choice procedure to circumvent this drawback is to compute $R_A := A^T A$ with dimension $M \times M$; then the first $M$ eigenvectors of the matrix $K_A$ can be expressed as a linear combination of eigenvectors $V$ of $R_A$ with the matrix of differences $A$. Therefore, the space of images $U$ can be generated so that $U:=A*V$, with dimension $NK \times M$, $U$ can be generated so that $U:=A*V$, with dimension $NK \times M$;

5) Build the training matrix $\Omega = U^T * A$ that represents the projection of $Db\_img$ into the image space, where each column of $\Omega$ corresponds to reduced image of $Db\_img$;

### C. Refining Step 3

The projection of the test fingerprint ($I_X$) within the space of images (U) occurs in the following chain:

$I_X \rightarrow$ vectorization $\rightarrow I_{XV} \rightarrow$ difference from the $mean(I_{Med}) \rightarrow$
$\rightarrow dI_{XV} = (I_{XV}(-)I_{med}) \rightarrow$ projection in the $U$ space $\rightarrow$
$\rightarrow PI_{XV} = U^T * (dI_{XV})$.

At this point we have now the fingerprint space ($U$), with all training vectors of the base ($\Omega$) and the image vector of test ($PI_{XV}$).

### D. Refining Step 4

Define a criterion for classifying fingerprints: Most strategies of image classification use Euclidean/Mahalanobis distance [11] for defining classification thresholds. The Euclidean distance is a spatial distance measure, whereas the Mahalanobis distance is a similarity measure.

1) Calculate the Euclidean distances $d_E(\Omega_j, \Omega_k)$ between all pair of vectors of the training matrix according to:

$$\Omega = \begin{bmatrix} \Omega_{11} & \Omega_{12} & \Omega_{1M} \\ \Omega_{21} & \Omega_{22} & \Omega_{2M} \\ \vdots & \vdots & \vdots \\ \Omega_{M1} & \Omega_{M2} & \Omega_{MM} \end{bmatrix} \rightarrow \Omega_j := \begin{bmatrix} \Omega_{1j} \\ \Omega_{2j} \\ \vdots \\ \Omega_{Mj} \end{bmatrix} \text{ and } \Omega_k = \begin{bmatrix} \Omega_{1k} \\ \Omega_{2k} \\ \vdots \\ \Omega_{Mk} \end{bmatrix}$$

$$d_E(\Omega_j, \Omega_k) := \sqrt{\sum_{z=1}^{n}(\Omega_{zj} - \Omega_{zk})^2}.$$

2) Calculate the Mahalanobis distances $d_M(I_{XV}, \Omega_k)$ between the fingerprint vector under test ($PI_{XV}$) and vectors $\Omega_{1,2\ldots M}$ of the training matrix. $d_M(I_{XV}, \Omega_k)$. For the framework of this algorithm, was adjusted and calculated as follows:

$$d_M(I_{XV}, \Omega_k) = \sqrt{(\Omega_k - PI_{XV})^T * [\lambda_{kk}]^{-1} * (\Omega_k - PI_{XV})},$$

where:

$$\Omega_k = \begin{bmatrix} \Omega_{1k} \\ \Omega_{2k} \\ \vdots \\ \Omega_{Mk} \end{bmatrix} \text{ and } PI_{XV} = \begin{bmatrix} PIx_{11} \\ PIx_{21} \\ \vdots \\ PIx_{M1} \end{bmatrix}$$

and $\lambda_{kk}$ are the elements of the diagonal matrix of eigenvalues corresponding to eigenvectors of the matrix $V$ (see Step B-4).

### E. Refining Step 5

Decide whether $I_X$ belongs or not to the fingerprint base. Set a threshold value $\theta_L := (1/2)\max[d_E(\Omega_j, \Omega_k)]$.

1) First case: if $\min[d_E(PI_{XV},\Omega)] \leq \theta_L$ then the test fingerprint is classified as belonging to the base;
2) Second case: if $\min[d_M(PI_{XV},\Omega)] \leq \alpha\lambda_{11}$ then, the acquired fingerprint is classified as belonging to the base, where $\alpha$ is factor for threshold adjustment and $\lambda_{11}$ is the largest eigenvalue as previously defined;
3) Third case: if $\min[d_E(PI_{XV},\Omega)] \leq \beta\lambda_{11}$ then, the test image is classified as belonging to the base, where $\beta$ is an adjustment factor of the threshold.

These criteria are frequently found in the literature and widely applied in face recognition schemes [7-9].

### III. THE H PARAMETER

Consider the requirement of finding an appropriate threshold to be used in this algorithm for inputs with fingerprint images. After a number of experiments with Euclidean and Mahalanobis distances, we identify a new parameter ($H$) and a new methodology to evaluate thresholds for the fingerprint classification. The parameter $H$ was experimentally defined as

$$H := \frac{[\min(d_M(PI_{XV},\Omega))]^2}{\min[d_E(PI_{XV},\Omega)]}$$

Intuitively $H$ is a brief of the second and third criteria described in Section II-E.

#### A. Methodology for assessing thresholds for classification

The choice of a value for the parameter $H$ can be done according to the following procedure:
1) Split the base Db_img into two sets, such that all elements of the first belong to the base and the all the elements of the second set are out of the fingerprint base;
2) Run the algorithm FCP by taking all images in Db_img as test image;
3) Calculate the value of $H$ for each test fingerprint image;
4) Generate a chart $H \times$ Image index($n$);
5) Get information from the graph about the values of $H$ to be used as decision thresholds.

This Methodology was applied for three different bases (DB3_B, DB2_B, DB1_B), which are available in the 'Fingerprint Verification Competition' (FVC) website [19]. Each DB is made up of eight different variations of 10 fingerprints, which results in eighty images per base. Each image is an 8-bit gray scale image stored in the TIFF format.

Fig 1 shows the plot of ($H \times I_X$) for the case of the base DB3: (1a) shows the response of $H$ in a noiseless scenario, while (1b) shows the outcome in the case of the presence of noise Gaussian (0.01, 0.1), with MATLAB[TM] 7.6.0. These graphs show an apparent separation between the regions with images inside and outside the data base. We can choose, from these plots, suitable $H$ values that minimize the occurrence of FP and FN. In this particular case the range $0.5 \leq H \leq 0.55$ was assumed as inconclusive, $H \leq 0.5$ the tested fingerprint is within the image bank, and $H \geq 0.55$ the tested image is assumed as is not belonging to the fingerprint bank. However, in the case of the fingerprint bases DB2_B DB1_B, as shown in Fig 2(a) and 3(a), as well as in the case of Figs 2(b) and 3(b) where Gaussian noise was added is not possible to settle a "separation region" due to the noisy nature of the fingerprints. Thus, it is not clear how to set suitable values for $H$ that control the occurrence of false positives and false negatives.

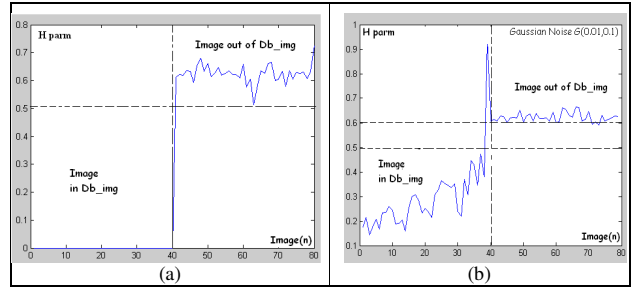

Fig 1: (a) and (b) show the disconnection between regions with images inside and outside the fingerprint base, which allows a suitable choice of a threshold for $H$ to control the rate of false positives and false negatives. The scenarios for which the curves (a) and (b) have been generated differ only by the presence of noise in Fig. 1(b).

Although we have considered the contamination of the digitalized fingerprint with additive noise trying to simulate more real cases, results show that the presence of noise can narrow the separation region, thus limiting the use of the parameter $H$. In the following section we present an issue that allows the disconnection of regions even under noisy conditions.

### IV. INCLUDE EDGE DETECTION IN FCP

As discussed at the end of the previous section, depending on the noise level, no clear separation can be found, thereby restricting the use of the parameter $H$. In this section we show that the use of type Canny edge or Sobel edge detection [10] can help mitigating the influence of noise. The edge detection is applied at the moment that the image is inserted into the base and when the test image is acquired. With the inclusion of edge detection component the complete sequence of the algorithm is:

Step 1: Store previously the fingerprint database denoted by *Db_img* (Fingerprint databank);
Step2: Apply Edge Detections:
Step3: Build the image space by using the PCA technique;
Step4: Define the criteria for classification from the H-curve;
Step5: Acquire, apply edge detection and project the fingerprint testing image ($I_X$) into image space;
Step6: Decide whether $I_X$ belongs or not to the stored base.

We apply "Canny" edge detection component of the MATLAB[TM], with DB2 and DB1 image bases. The results can be seen in the in Fig 4, 5 and 6. Fig 4 and 5 show the "$H$-plot" with DB2 for combinations with and without edge detection.

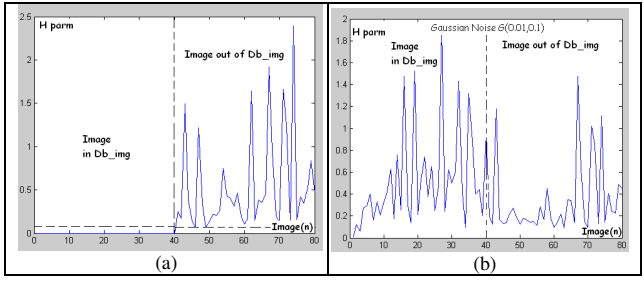

Fig 2: DB2_B (a) the separation between the regions, even without noise occurs in a narrow band around $H = 0$. (b) region of separation does not occur, and the values of $H$ drastically increase the potential of false positives and negatives.

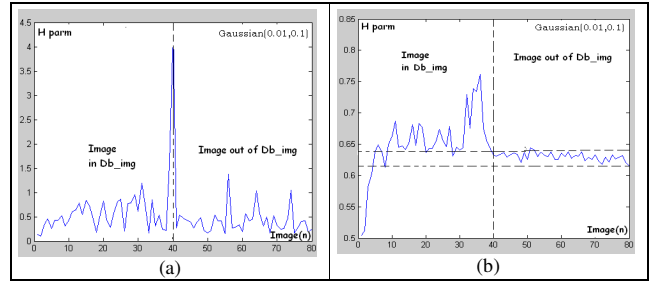

Fig 5: DB2 Fig. 5(a) shows the fingerprint classification with noise, but without edge detecting. Fig. 5 (b) shows the same case, but considering the edge detecting.

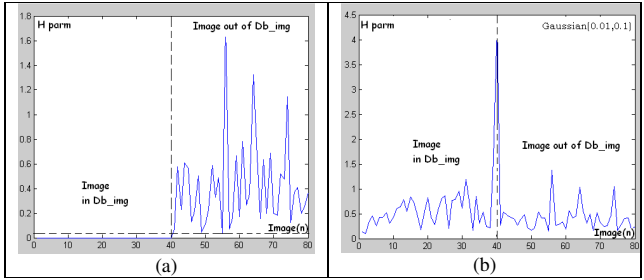

Fig 3: DB1_B (a) the separation between the regions, even without noise occurs in a narrow band around $H = 0$. (b) region of separation does not occur and the values of $H$ increase the potential of false positives and negatives.

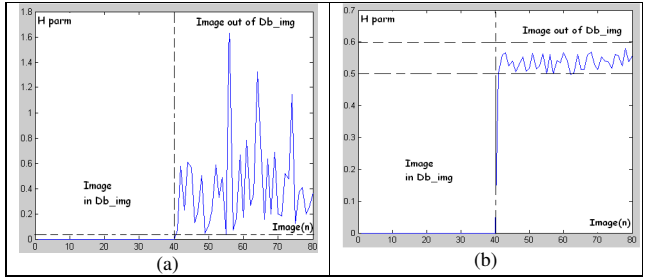

Fig 6: DB1 Fig. 6(a) shows the case without noise and without detect edge. Fig. 6(b) shows the noiseless fingerprints, but now using edge detection before processing.

. Fig 4(a) shows the case without noise and without edge detection. Fig 4(b) shows the noiseless case, but with edge detection. It is worth to remark that the inserting of edge detection grants a wide separation between the regions so the $H$ parameter can be chosen to control the occurrence of FP and FN. Fig 5(a) shows the case with noise and without edge detecting. Fig 5(b) shows the fingerprint analysis with both noise and edge detection. It can be remarked that with edge detection there is a slight separation between regions, just enough for a suitable choice of $H$ that meets also the case without noise shown in Fig 5(b). In this fingerprint analysis we set up the following conditions:

if $0.6 < H \leq 0.64$ then $I_x$ is assumed to be out of the base.

Fig 6(a) shows a noiseless fingerprint classification without involving edge detecting. Fig 6(b) shows a FPC case with noise, but with edge detecting. We note that the application of edge detection in this case, affords a wider separation between the regions and $H$ can be chosen to drive the tradeoff between the occurrence of FP and FN in the fingerprint classification. In this case we use Sobel edge detection [10], but there was no significant improve. At this point we have the full algorithm. In the next section the ROC curve is used to assess the overall performance of the FCP.

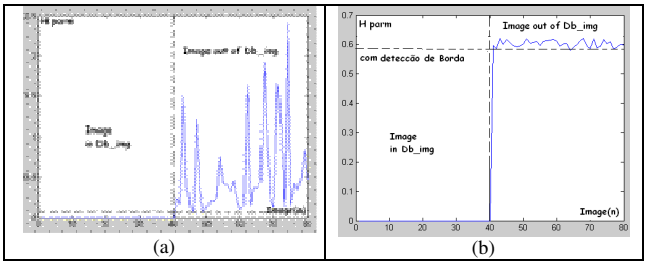

Fig 4: DB2 Fig. 4(a) shows the case without noise and without detecting edges. Fig. 4(b) shows the noiseless case, but now including edge detection.

## V. PERFORMANCE ANALYSIS USING ROC

ROC curve is a graphical plot related in a natural way to cost/benefit analysis of binary classifier systems as its discriminant threshold is varied [6]. In this section the ROC is derived to assess the overall performance of the algorithm for the three databases (DB1, DB2 and DB3), through accounting of False Positives (FP) and False Negatives (FN). The frequency of occurrence of FN and FP is strongly noise-level dependent and also may depend upon the size of the database of images projected into image space. Therefore the evaluation scenario was narrowed as follows: defining four noise levels (high, medium, low, no noise) for fingerprint images of the databases BD1, BD2 and BD3. Such additive Gaussian noises are generated and added by the built-in function imnoise ($I$, 'gaussian', $m$, $v$) of Matlab[TM]. Fig. 7 shows the ROC curves for DB1 database. The ROC for databases was computed using forty training fingerprint images. Four different noise levels are considered, and the $H$-parameter was varied from zero to unity. Each point of the curve corresponds to a particular value of $H$. The DB1-ROC suffers a huge variation in the rate (FN, FP), as shown in Fig 7, In this case, the noiseless performance achieves (0.0), and the insertion of low or medium noise yields (0.3,0,3), jumping abruptly with the insertion of high noise.

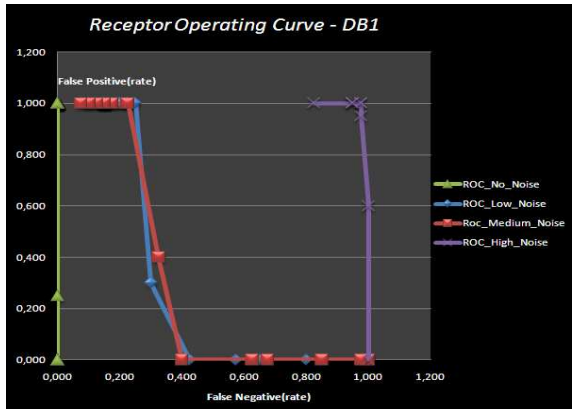

Fig 7: ROC curve for the fingerprint database DB1.

The tradeoffs between FN and FP are better shown in Figures 8 and 9 in the case of medium noise. Figure 8 shows that a performance around (0.0) can be obtained by growing the DB2 database size. As a general rule, there is a tradeoff between the rates (FN, FP), that is, decreasing one rate increases the other, as expected. A choice of *H* to achieve suitable rates can be done even for noisy fingerprint images. Another relevant aspect is that each case is unique, i.e. each fingerprint database both size (Fig.8) and quality requires adapting the methodology to establish the parameter *H*.

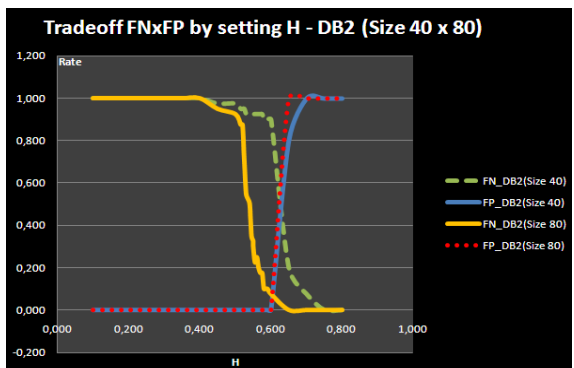

Fig 8: Tradeoff (FN, FP) to DB2 size (40 × 80). In this case we can get rates ranging from (0.4, 0.4) *H*=0.7 to (0.05, 0.05) *H*=0.6.

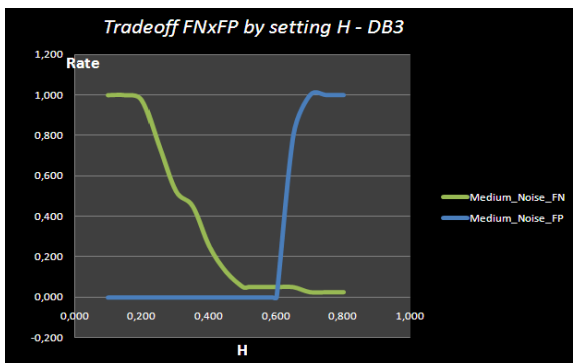

Fig 9: Tradeoff (FN, FP) for the database DB3(size 40). We can set *H*=0.60 to obtain a performance (FN, FP) around (0.0).

## VI. CONCLUDING REMARKS

Fingerprint image has long been used in applications involving biomedical-based authentication solutions, forensic analysis or in environments that require access control. The strategy proposed in this paper is not intended to replace existent on. Depending on the purpose the facilities where the access is to be controlled, acceptable rejection rates (false negative) and error (false positive) may be different. This can be exploited in the ROC. The picture here can be far from FBI recommendations for fingerprint recognition (accuracy≥99%, rejection≤20%). High-security sites prefer to guarantee low error rates, and then minimize the rejection as much as possible. In contrast, less secure places may prefer to ensure certain low rejection level and then minimize the error rate. This approach can also be used as a preliminary screening test as a first step in a recognition system (fingerprint matcher). If the algorithm points out that the acquired fingerprint is within a very large database, then a second and more sophisticated retrieval algorithm is applied.